\documentclass[11pt,letterpaper]{article}
\usepackage{ijcnlp2017}
\usepackage{times}
\usepackage{latexsym}

\def\ijcnlppaperid{147}

\usepackage{color,soul}
\usepackage{caption}
\usepackage{multirow}
\usepackage{hyperref}
\usepackage{cleveref}
\usepackage{amsmath}
\usepackage{amssymb}
\usepackage{graphicx}
\usepackage{array}
\usepackage{tabularx}
\usepackage{caption}
\usepackage{mathptmx}
\usepackage{subfigure}
\usepackage{float}
\usepackage[utf8]{inputenc}

\usepackage{graphicx,calc}
\newlength\myheight
\newlength\mydepth
\settototalheight\myheight{Xygp}
\newcommand*\inlinegraphics[1]{%
  \settototalheight\myheight{Xygp}%
  \settodepth\mydepth{Xygp}%
  \raisebox{-\mydepth}{\includegraphics[height=\myheight]{#1}}%
}

\ijcnlpfinalcopy

\title{Cross-Lingual Sentiment Analysis Without (Good) Translation}

\author{Mohamed Abdalla \and Graeme Hirst \\
  {\tt msa@} \and {\tt gh@cs.toronto.edu}
  \\ Department of Computer Science, University of Toronto \\ 
  Toronto, Canada}

\date{}
\graphicspath{{./figs/}}
\begin{document}

\maketitle

\begin{abstract}

Current approaches to cross-lingual sentiment analysis try to leverage the wealth of labeled English data using bilingual lexicons, bilingual vector space embeddings, or machine translation systems. Here we show that it is possible to use a single linear transformation, with as few as 2000 word pairs, to capture fine-grained sentiment relationships between words in a cross-lingual setting. We apply these cross-lingual sentiment models to a diverse set of tasks to demonstrate their functionality in a non-English context. By effectively leveraging English sentiment knowledge without the need for accurate translation, we can analyze and extract features from other languages with scarce data at a very low cost, thus making sentiment and related analyses for many languages inexpensive.

\end{abstract}

\section{Introduction}

Methods for sentiment analysis and classification have largely been limited to English, making use of large amounts of labeled data to produce sentiment classification. As a consequence, many developed approaches cannot be readily applied to other languages, which usually do not have the wealth of labeled data that is exclusive to English. Therefore many approaches which deal with other languages often:  i) experiment with small datasets that are limited in domain or size of training and testing sets \citep{P11-1001, tan:08}, or ii) attempt to elucidate sentiment lexicons for their respective languages \citep{P16-1001}.

A growing number of publications attempt to leverage labeled English data to compensate for the relative lack of training material in the other languages. This is usually done through the use of either bilingual lexicons \cite{P12-1001}, machine translation (MT) systems \citep{P15-1001, Zhou:16}, or more recently, through the use of bilingual vector space embeddings \citep{Chen:16}.

Unfortunately, in some cases such data is still expensive to obtain. Many languages do not have good, or sometimes any, MT systems, and the cost of producing word alignments or sentence alignments for training bilingual word embeddings (BWE) \cite{P13-1001, Bengio:15} or similar techniques \citep{P15-1002} is prohibitive for data-poor languages.

Here we introduce a high-performance, low-cost approach to cross-lingual sentiment classification, which can be used to benchmark more expensive methods. We demonstrate the utility of this approach by highlighting how very limited training data suffices for effective cross-lingual sentiment analysis in various contexts (both at the word and sentence/document level). Our approach relies on the simple vector space translation matrix method \citep{Mikolov:13}, which computes a matrix to convert from the vector space of one language to that of another. It hinges on the observation that sentiment is highly “preserved” even in the face of poor translation accuracy. We observed that a sentiment classifier trained only with word vectors from English (hereafter referred to as the target language) performs well on unseen words from other unseen languages (referred to as the source languages) that are translated into the English vector space through the simple matrix method, even with very poor translation scores.\footnote{This work involves transfers in both directions between English and other languages. We will take the perspective of the translation matrix (section 3.2) and refer to English as the target language and the others as source languages.}

We quantitatively evaluate our methods by training on English and testing on words from Spanish and Chinese. We chose these languages in order to show: i) the robustness of the technique irrespective of the grammar of the source and target languages and, ii) the ability to validate the technique ({\em i.e.,} as baselines for experiments).  We emphasize that, regardless of the source languages used, we treated them as if they had no MT systems. We made use of only 2000, 4500, or 8500 word pairs -- a task easily accomplished by a human translator on a data-poor language. We experiment with differing amounts of data during training to show the robustness of our observation. We then apply the fine-grained sentiment regressor to the task of review classification as done by \citet{Chen:16}, and show that our na{\"{i}}ve algorithm achieves results similar to their benchmark but at a much lower cost.

\section{Previous Work}

\subsection{Cross-Lingual Word Embeddings }
Monolingual word embedding algorithms use large unlabeled datasets to learn useful features about the given language \cite{pennington:14, mikolov:2013d}. These algorithms learn vector representations for the words of the language --- an encoding that has proven utility in a variety of NLP tasks including sentiment analysis \cite{maas:11} and machine translation \cite{Mikolov:13}.
 
When working with more than one language, we seek to satisfy two objectives: i) monolingually, similar words of the same language have similar embeddings; and ii) cross-lingually, similar words across languages also have similar embeddings. Satisfying these two criteria would allow us to use algorithms trained for the embeddings of a single language (such as English, with a wealth of labeled data) for other languages as well. Below we discuss algorithms to achieve the cross-lingual objective, their costs, performance, and the rationale underlying our algorithm design.

\subsubsection{Offline Alignment}

The simplest approach to achieving the cross-lingual objective is to train each monolingual objective separately (create a model for each language), and then learn a transformation to enforce the second objective. This approach uses a dictionary of paired words in order to learn a transformation or `alignment' from the vector space of one language to that of another. 

First introduced by \citet{Mikolov:13}, and later extended by \citet{faruqui:2014}, this offline alignment is fast and low cost, but does not achieve a high translation accuracy. A big drawback of these approaches is that using a dictionary ignores the polysemic nature of languages. It is also not clear or proven that a single transformation would be able to capture the relationship between all the words in a cross-lingual setting.

We opt to use offline alignment to show that such a low-cost approach does, in fact, capture a significant part of the relationship between words of different languages when it comes to sentiment. That is, a single transformation (linear in the case of our work) is sufficient to learn a projection which allows one to use labeled English data to aid in sentiment analysis.

\subsubsection{Parallel-Only}
An alternative approach to offline alignment is the parallel-only approach. Approaches which fall into this group, such as BiCVM \citep{hermann:2013} and bilingual auto-encoder (BAE) \citep{ap:2014}, rely exclusively on sentence-aligned parallel data to train a model with similar representations. Such approaches can be effective, but require extremely expensive data. Another drawback is that these approaches can be affected by the writing style of the parallel text \cite{Bengio:15}.

\subsubsection{Jointly-Trained Model}

Combining the offline alignment and parallel-only algorithms is a third class of jointly-trained approaches. These approaches jointly optimize the monolingual objective at the same time as the cross-lingual objective, making use of both monolingual and parallel data. Approaches like those of \citet{klementiev:2012} and \citet{P13-1001} use word-aligned data in order to learn the fine-grained cross-lingual features and tend to be quite slow. Other approaches (including that of \citet{Bengio:15}) rely on sentence-aligned data and are faster than those using word-aligned data. While these models are more cost-efficient than parallel-only approaches, it remains expensive and sometimes prohibitive to obtain sentence alignments for many languages (a problem that we seek to avoid).

A lower-cost alternative to these expensive jointly trained models was proposed by \citet{Duong:16} and later used to project multiple languages in the same vector space \citep{Duong:17}. The model involved creating and making use of translations produced using a bilingual dictionary during training. Using expectation-maximization--inspired training, sentence translations were produced by selecting translations of words based on context to deal with polysemy, and this approach demonstrated improvements on the simple linear transformation method. However, even this model uses a significantly larger amount of data than the methods used in this work, with its smallest dictionary being composed of 35,000 word pairs compared to our approach, which can use as little as 2000 words  for both translation and sentiment regression.

\subsection{Cross-Lingual Sentiment Analysis}

Previous approaches to cross-lingual sentiment analysis can be classified into two main categories: i) those that rely on parallel corpora to train BWE's ({\em i.e.,} they use pre-trained embeddings) \citep{Chen:16, ap:2014, tang:2014}, and ii) those that use translation systems \citep{zhou:15, Zhou:16} in order to obtain aligned inputs to learn to extract features which work on both languages. Both approaches allow the sentiment portion of training and testing data to be in the same vector space. However, many languages have no MT system, and it is extremely expensive to create one on a language-by-language basis. 

Our proposed approach is simpler in that it requires only a small word-list to learn both the embedding and the sentiment classification, the duality of which cannot be claimed by previous approaches.

\section{Methods and Data}

\subsection{Data}

\subsubsection{Vector Space Data}

{\bf English Vector Space Model} For English we used a model pre-trained on part of the Google News dataset (which is composed of approximately 100 billion words).\footnote{https://code.google.com/archive/p/word2vec/} The words are represented by 300-dimensional vectors.

\noindent {\bf Spanish Vector Space Model} For the Spanish word embeddings we opted to use a model pre-trained on the Spanish Billion Word Corpus \cite{card:16}. It consists of just under $1.5 $ billion words compiled from a variety of Spanish resources. As with the English model, the words are represented by 300-dimensional vectors.

\noindent {\bf Chinese Vector Space Model} For Chinese word embeddings we learned our own vector representations using a Wikimedia dump\footnote{https://dumps.wikimedia.org/zhwiki/latest/} composed of around $150$ million words from $250,000$ articles in both simplified and traditional Chinese. We used OpenCC\footnote{https://github.com/BYVoid/OpenCC} to translate the articles in simplified Chinese to traditional. To segment the text into tokens we used Jieba\footnote{https://github.com/fxsjy/Jieba}. Finally, to create the actual word embedding model, we used Gensim \cite{vre:11} with the minimum count set to 1, using continuous bag of words (CBOW), a window of 8, and vector dimension set to 300. 

\subsubsection{Word Lists}

{\bf Translation Word List} For the process of learning a translation matrix from one language to the other, a lexicon of approximately $10,000$ English words was obtained online by scraping the most commonly used words as determined by n-gram frequency analysis in Google's ``Trillion Word Corpus"\footnote{https://github.com/first20hours/google-10000-english}. The lexicon was then translated using Google Translate\footnote{https://translate.google.com/} in order to obtain corresponding words in Spanish and Chinese. For alignment lists of smaller sizes during experimentation, a random subset of the larger list was selected. During the randomized selection, we discarded any words which were not in the target language vector space and whose translation was not in the source language(s) vector space(s).

\noindent {\bf Binary Sentiment Word List} For the task of binary sentiment classification we used a list\footnote{ https://github.com/williamgunn/SciSentiment} curated by \citet{hu:2004} containing both positive and negative  English opinion words (or sentiment words). Google Translate was used to translate the list into the other languages to obtain cross-lingual word pairs. During training and testing we made sure to balance the dataset and to discard words that were not in the vector space of the target language or whose translation was not in the vector space of the source language(s).

\subsubsection{Fine-Grained Data}
For fine-grained sentiment regression, we used Affective Norms for English Words (ANEW) \cite{bradley:1999a}. The creators of ANEW sought to provide emotional ratings for a large number of words in the English language.

ANEW proposes that all human emotion can be organized in a vector space with three basic underlying dimensions (or axes). The first dimension, {\em valence}, ranges from {\em pleasant} to {\em unpleasant}; the second dimension, {\em arousal}, ranges from {\em calm} to {\em excited}; and the third dimension, {\em dominance}, ranges from {\em in-control} to {\em out-of-control}. Examples are shown in Table \ref{table:ANEW}.

\citet{bradley:1999a} used a nonverbal pictographic measure, the Self-Assessment Manikin (SAM) \citep{bradely:94}, to measure stimuli across these three dimensions. The figures in the SAM consist of bipolar scales depicting different values along each of the three emotional dimensions. For example, when considering valence, SAM ranges from a frowning unhappy figure to a smiling happy figure. Similar ranges are extended across the two other dimensions. Using this test, Bradley and Lang were able to arrive at a numerical value representing a word's stimulus for each dimension ranging from 1 to 9; where 1 is the low value ({\em unpleasant, calm, in-control}) and 9 is the high value ({\em pleasant, excited, out-of-control}).

\begin{table}[]
\centering

\begin{tabular}{rll}
          & \textbf{Low Stimulus} & \textbf{High Stimulus} \\
\textbf{Arousal}   & {\em relaxed} (2.39)        & {\em infatuation} (7.02)     \\
\textbf{Dominance} & {\em victim} (2.69)         & {\em confident} (7.68)       \\
\textbf{Valence}   & {\em death} (1.61)          & {\em beauty} (7.82)         
\end{tabular}
\caption{Examples of words on each end of the spectrum for each of the three dimensions of ANEW. Numeric stimulus value is shown in parentheses.}
\label{table:ANEW}

\end{table}

\subsubsection{Review Data}

In this subsection, we discuss the data used to replicate the review classification task done by \citet{Chen:16} as a means of validating the utility of our model. This experiment was done using only English and Chinese because there was no Spanish data for this task and the Arabic review data-set that \citet{Chen:16} used was not freely available.

\noindent {\bf Labeled English Reviews} Following \citet{Chen:16}, we obtained a balanced dataset of $700,000$ reviews of businesses on Yelp from \citet{zhang:15c} with their sentiment ratings as labels ranging from 1 for very negative to 5 for very positive.

\noindent {\bf Labeled Chinese Reviews} Here we use a dataset from \citet{lin:15}. Their work  provides hotel reviews, with labels ranging from 1 for very negative to 5 for very positive. In order to fairly compare our work with that of \citet{Chen:16}, we use $10,000$ reviews for model selection, and another unseen $10,000$ as our test set.

\subsection{Translation Matrix Technique}

As described by \citet{Mikolov:13}, the translation matrix technique assumes that we are given a set of word pairs and their associated vector space representations. More specifically,  we are given $j$ word pairs, $\{x_{i}, z_{i}\}_{i=1}^{j}$ where $x_i \in \mathbb{R}^n $ is a vector from the source language of word $i$ and $z_{i} \in  \mathbb{R}^{m}$ is the vector representation  of the corresponding translated word in the target language.

We then want to find a transformation matrix $W$ such that $W x_{i}$ approximates $z_{i}$. We learn this by solving the following optimization problem: 
\[ \min_{W} \sum_{i=1}^{j} ||W x_{i} - z_{i}||^2 \]
Instead of solving with stochastic gradient descent, we instead opt to use the closed-form solution.

To translate a word from source language to target language, we can map it using $z = W x$, and then find the closest word in that language space using cosine similarity as the measure of distance. The method of testing was Monte Carlo cross-validation run 10 times with a split of 90\% training data and 10\% test data.

\subsection{Models}

\begin{figure}[t]
\centering     

\includegraphics[width=0.5\textwidth]{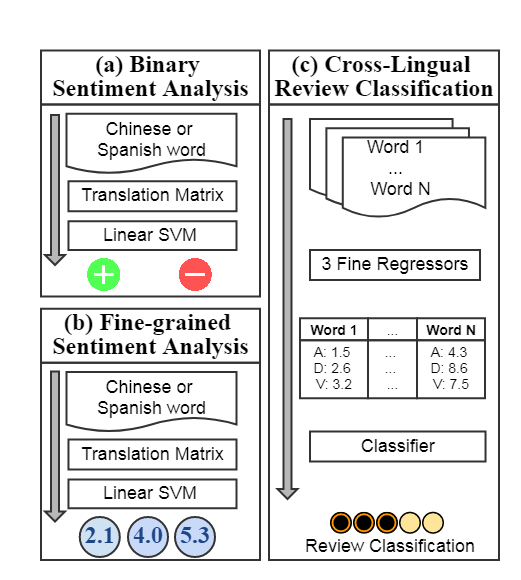}

\caption{The three models used in the experiments.}
\label{fig:models}
\end{figure}

\subsubsection{Binary Sentiment Analysis Model}

The sentiment analysis model, Figure \ref{fig:models}(a), is a simple linear support vector machine (SVM) classifier. Implemented using Sci-kit Learn's {\em SGDClassifier} function \cite{Pedregosa:11}, this model takes a word represented as a vector with dimension of 300 and outputs a prediction of either $-1$ or $+1$ (for negative and positive respectively). The classifier itself performs stochastic gradient descent (SGD) with $l2$ regularization to arrive at the best classification.

The training procedure of this model is quite simple and involves only the target language. The model is trained only on word embeddings from the target language but tested on embeddings returned by the translation of words originally from the source language(s). We made sure that the English translation of the test words had not been seen before in training. The training/testing split was changed to 80\% and 20\% from the previous 90\%/10\% to account for the smaller number of examples in the dataset (and the fact that we wanted to test on a representative sample of the data).

\subsubsection{Fine-Grained Sentiment Analysis Model}

The fine-grained sentiment analysis model in Figure \ref{fig:models}(b) is a regression model to predict the ANEW values for each of the three dimensions. For this task, we built a regressor for each of the three dimensions whose input is a 300-dimension vector and whose output is a real number from 1 to 9. The regressor used was a Bayesian Ridge regressor, which estimates a probabilistic model of the regression problem. The prior for the parameter $w$ is given by a spherical Gaussian: \[p(w|\lambda) = \mathcal{N}(w|0,\lambda^{-1}I_p) .\]

\noindent The model is similar to that of the Ridge regression. The model was implemented using Sci-kit Learn's with hyperparameters alpha\_1 and alpha\_2 set to 1.

The training procedure of this model is quite simple and similar to that of the previous model. Again, it only trains on the vector of the target language, and is tested purely on words from the source language whose translation into the target language was not seen in the training of the regressor (so that there may be no chance of skewing the results). The training/testing split here was 75\%/25\%.

\subsubsection{Review Classification Model}

For the task of review classification, another model Figure \ref{fig:models}(c), was built to make use of the previously described fine-grained sentiment analysis model. The classifier used for this task is a logistic regression classifier. For a given review $r_{i}$ in the target language, composed of $n$ words, we construct a $1 \times \mathit{Max\_length}$ sentiment vector (where $\mathit{Max\_length}$ is the number of words of the longest target review). We construct this vector by passing in the word embedding for every word into the sentiment analysis model and placing the resulting values (of $1$ to $9$) into the constructed array. The array is then padded with $0$'s in order to make it of length $\mathit{Max\_length}$.

For reviews in the source language, the process is similar, with the only change being that the words are first translated from their original vector space to that of the target language before being passed into the sentiment classifier. Once the review vector has been constructed, it is passed to the classifier to produce a classification of 1 to 5.

As with the previous classifier, the training procedure is quite simple and involves only the target language ({\em i.e.}, the classifier is trained only on reviews which are originally from the target language, but it is tested on reviews only from the source language).

\section{Results}

\subsection{Translation Accuracy}

\begin{figure}[t]
\centering     

\includegraphics[width=0.5\textwidth]{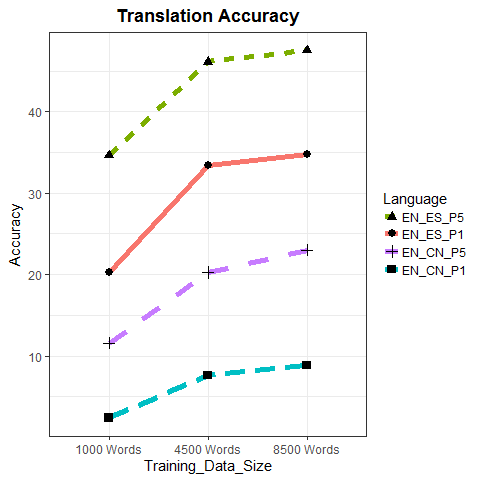}

\caption{Top-1 and Top-5 accuracy for Spanish and Chinese using various amounts of training data.}
\label{translationacc}
\end{figure}

\begin{table}[]
\centering
\captionsetup{}

\resizebox{\columnwidth}{!}{
\begin{tabular}{lrrrrrr}
                     & \multicolumn{2}{c}{\textbf{1000 WORDS}} & \multicolumn{2}{c}{\textbf{4500 WORDS}} & \multicolumn{2}{c}{\textbf{8500 WORDS}} \\
\textbf{Translation} & P@1                & P@5                & P@1                & P@5                & P@1                & P@5                \\
EN $\rightarrow$ ES  & 20.3               & 34.6               & 33.42              & 46.13              & 34.79              & 47.79              \\
EN $\rightarrow$ CN  & 2.4                & 11.6               & 7.60                & 20.29              & 8.87               & 23.01             
\end{tabular}
}
\caption{Accuracy of the word translation method. P@1 and P@5 represent Top-1 and Top-5 accuracy respectively.}\label{table:table1}
\end{table}

We first measure the accuracy of the matrix translation method using the same test described by \citet{Mikolov:13}. The purpose of these tests is two-fold: i) verifying that the data used and implementation of the technique reproduces what is expected, and ii) quantifying how much sentiment is preserved with low translation accuracy (by explicitly noting the poor accuracy of the translation).  The data used is described in section 2.1. Table \ref{table:table1} shows the effect of training size (number of words) on the accuracy of the translation matrix method. As expected (and previously shown by \citet{Mikolov:13}), the translation accuracy increases with more training examples, as shown in Figure \ref{translationacc}. The data and methods used in this work are further validated as the translation accuracy results closely approximate those of \citet{Mikolov:13}, with English-Spanish translational accuracy achieving 35\% and 48\% accuracy for P@1 and P@5 respectively when compared to the original 33\% and 51\%. This concordance is further validation of the method. It is also interesting to note that Chinese, which is less like English than Spanish is, also suffers a lower translation score across both categories. However, we see that this large drop is not represented significantly in later portions.

\subsection{Binary Word Sentiment Classification}

\begin{figure}
\centering     
\includegraphics[width=0.5\textwidth]{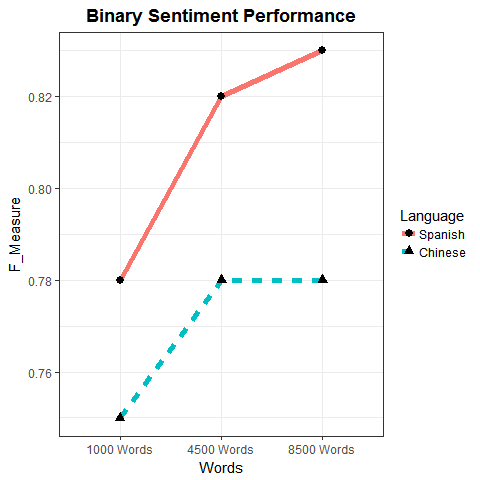}
\caption{F-measure of binary sentiment classifier with varying amounts of training data for both Spanish and Chinese.}
\label{fig:binaryacc}
\end{figure}

The second experiment tested the binary cross-lingual sentiment classification capabilities of the matrix translation method, {\em i.e.}, how well can we differentiate between positive and negative words of a language we have not seen before using a model trained only on English words? Here we used the binary sentiment word list described in section 2.1.2 in order to assess whether or not the translation matrix would preserve sentiment even with poor translation accuracy scores. Given that the classifier is trained only on the target language vectors, we used the translation matrices produced previously to translate a word from source to target language embedding space.

As we can see in Table \ref{table:table3} and Figure \ref{fig:binaryacc}, even with low translation accuracy, such as the 1000-word Chinese translation matrix, we are able to achieve good binary sentiment classification. We also noted that a significant drop in translation accuracy results in only a relatively small drop in sentiment classification performance.

\begin{table}[t]
\centering
\captionsetup{}

\begin{tabular}{lcc}
          & \textbf{Spanish} & \textbf{Chinese} \\
\multicolumn{3}{l}{\textbf{1000 Words}}         \\
Precision & 0.77             & 0.76             \\
Recall    & 0.79             & 0.74             \\
F-measure & 0.78             & 0.75             \\ \\
\multicolumn{3}{l}{\textbf{4500 Words}}         \\
Precision & 0.82             & 0.79             \\
Recall    & 0.82             & 0.77             \\
F-measure & 0.82             & 0.78             \\ \\
\multicolumn{3}{l}{\textbf{8500 Words}}         \\
Precision & 0.83             & 0.80              \\
Recall    & 0.83             & 0.77             \\
F-measure & 0.83             & 0.78            
\end{tabular}
\caption{Results of the binary sentiment classification task for each language with each translation matrix.} \label{table:table3}
\end{table}

\subsection{Fine-Grained Sentiment Analysis}

\begin{figure*}
\centering     
\subfigure[Arousal]{\label{fig:d}\includegraphics[width=50mm]{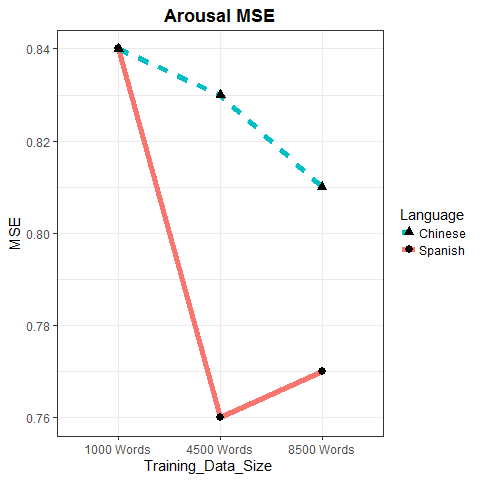}}
\subfigure[Dominance]{\label{fig:e}\includegraphics[width=50mm]{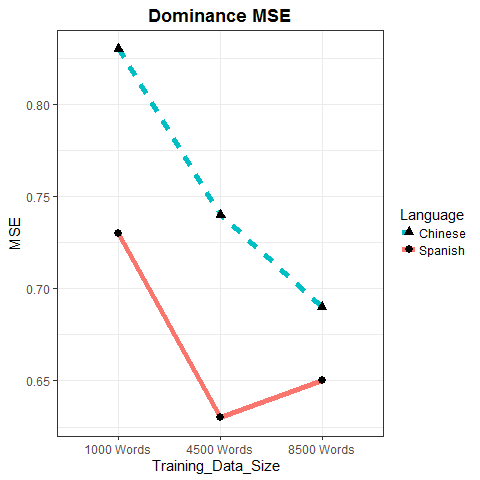}}
\subfigure[Valence]{\label{fig:f}\includegraphics[width=50mm]{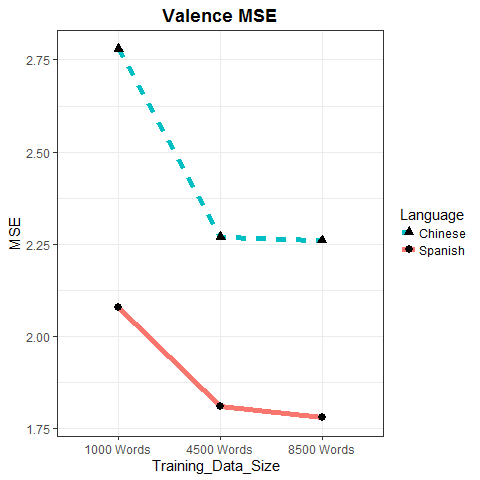}}
\caption{Mean squared error for each regression problem with varying amounts of training data for all three of the ANEW dimensions. (Lower is better).}
\label{fig:MSE}
\end{figure*}

In the third experiment, we tested the accuracy of cross-lingual regression when it comes to predicting a word's value in any of the three ANEW dimensions of {\em valence}, {\em arousal}, and {\em dominance}. We attempted to predict the {\em valence}, {\em arousal}, and {\em dominance} of words in source language, having only trained on target language. As we saw in the last experiment, massive drop-off in translation accuracy need not result in a massive drop-off in sentiment analysis. As this is a regression problem, Table \ref{table:table2} presents both the $r^2$ and the mean squared error (MSE) as measurements of model performance. The MSE per dimension is visualized in Figure \ref{fig:MSE}. Given the data's scale from 1 to 9 with an average standard deviation among participants for each word of 2.02, an average mean squared error of approximately 1 shows that our model has high predictive power.

\begin{table}[t]
\centering
\captionsetup{}

\begin{tabular}{lcc}
          & \textbf{Spanish} & \textbf{Chinese} \\
\multicolumn{3}{l}{\textbf{1000 Words}}         \\
Arousal   & 0.24 (0.84)      & 0.24 (0.84)      \\
Dominance & 0.31 (0.73)      & 0.23 (0.83)      \\
Valence   & 0.48 (2.08)      & 0.32 (2.78)      \\ \\
\multicolumn{3}{l}{\textbf{4500 Words}}         \\
Arousal   & 0.33 (0.76)      & 0.28 (0.83)      \\
Dominance & 0.39 (0.63)      & 0.28 (0.74)      \\
Valence   & 0.54 (1.81)      & 0.44 (2.27)      \\ \\
\multicolumn{3}{l}{\textbf{8500 Words}}         \\
Arousal   & 0.33 (0.77)      & 0.29 (0.81)      \\
Dominance & 0.38 (0.65)      & 0.31 (0.69)      \\
Valence   & 0.54 (1.78)      & 0.43 (2.26)     \\
\end{tabular}
\caption{Results of the fine-grained sentiment regression task for each language with each translation matrix, in the form of $r^2 (MSE)$.}\label{table:table2}
\end{table}

\subsection{Sentiment Classification of Reviews}

In the fourth experiment, we sought to show that the regressor developed in section 3.3.2 could be used as a feature extractor in performing other tasks. To this end, we replicated the experiment by \citet{Chen:16}, who predicted hotel ratings from Chinese reviews using a model trained only on English restaurant reviews. 

\citet{Chen:16} had two baseline models which they beat with their new model: i) a logistic regression classifier (line 1 in Table \ref{table:table4}), and ii) a non-adversarial variation of adversarial deep averaging network (ADAN) (line 2), referred to as DAN (deep averaging network), which is one of the state-of-the-art neural models for sentiment classification. These were the only two models which did not make use of either labeled Chinese examples or an MT system, and therefore were chosen to serve as a fair comparison to our method. Both models use bilingual word embeddings as an input representation to map words from both languages into the same vector space. Our own model (line 3 in Table \ref{table:table4}) uses logistic regression on sentence arrays created by predicting their ANEW values for each dimension to predict review scores. 

Table \ref{table:table4} shows that we were able to closely match the accuracy of the baseline systems implemented by \citet{Chen:16} for Chinese reviews. These results demonstrate that our sentiment regressors encoded enough information into the sentence vectors to achieve similar results to the baseline models which took bilingual word embedding as input, and that the fine-grained sentiment model can be used to extract sentiment-based features for other tasks in languages where aligned data might be expensive to obtain.

\begin{table}[]
\centering
\begin{tabular}{ l r  }
 \textbf{Approach}                 & \textbf{Accuracy} \\ 
Logistic regression (BWE)               & 30.58\%                    \\ 
Deep averaging network (DAN)                               & 29.11\%                    \\ 
Logistic regression (ANEW)  & 28.05\%                    \\  
\end{tabular}
\caption{Model performance on sentiment extracted vectors versus previous approaches. Logistic regression on predicted sentiment (ANEW) values preformed similariy to both regression and DAN on BWEs. }\label{table:table4}
\end{table}

\section{Discussion and Future Work}
We have shown that the matrix translation method can be used to infer and predict cross-lingual sentiment. More notably we observed that: i) sentiment is preserved accurately even with sub-par translations, and ii) this low-cost approach also maintained fine-grained sentiment information between languages.

We further cemented these observations through a variety of experiments. The first experiment preformed was testing the translation accuracy of the method presented to verify validity of the algorithm. The second experiment, was binary word classification into either positive and negative for words in Chinese or Spanish, given a model that was trained only with English. With a translation P@5 as low as 11\%, our linear classifier was still able to predict with ~75\% precision and recall the polarity of a word's sentiment. The third experiment, in which we were able to predict a word’s position on the three-dimensional ANEW 9-point scale with an error margin of 1 (per dimension), lent further credibility to the validity and data-efficiency of our approach. Our fine-grained sentiment analysis with the ANEW scale is notable as it demonstrates how the algorithm works at the word level --- useful in building sentiment lexicons in an automated fashion at a very low cost and with little manual effort. Our last experiment, Chinese review classification, further highlighted the robustness of our model, by showing that vectors created using the regressor encoded enough sentiment information to match the baseline methods of passing in bilingual word embeddings to a trained model. We showed how competitive the proposed approach is when compared to much more expensive methods and that it can directly be applied to sentiment classification tasks for data-poor languages.

Throughout the experiments, we saw the general trend of reduced error and increased accuracy with more training data. However the increase in accuracy starts to diminish with around 8500 words. The root of this leveling of accuracy could be the inherent limitation of either the translation technique used or the classification or regression algorithms used or both.

A surprising finding was how well sentence vectors composed of ANEW values for each word performed when compared to the baselines of \citet{Chen:16}, Table \ref{table:table4}. Achieving similar results to the previous baseline using the same classification algorithm (logistic regression) means that the sentence vectors composed of ANEW values encode enough (and as much) information as the BWE, or that the classification algorithm used on the BWEs isn't strong enough to extract more meaningful relationships between the values. The truth is probably a mixture of both reasons. 

By choosing the number and type of languages selected we have shown that this observation holds across languages with different roots and different grammar systems, which is further re-affirmed by the fact that we train our models only on English data, but test purely on Chinese and Spanish.

\begin{table}[]
\centering

\resizebox{\columnwidth}{!}{
\begin{tabular}{llr}
\textbf{\begin{tabular}[c]{@{}l@{}}Chinese Word\\ (English)\end{tabular}} & \textbf{\begin{tabular}[c]{@{}l@{}}Computed \\ English Translation\end{tabular}} & \textbf{\begin{tabular}[c]{@{}l@{}}True Valence\\ (Predicted Valence)\end{tabular}} \\ \hline
\inlinegraphics{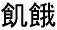} (Hungry)   & Mugabe misrule  & 3.58 (2.85) \\
    & Boo hoo & \\
    & Quagmire & \\
    & Chikwanine & \\
    & Sylvain Angerlotte & \\
\inlinegraphics{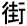} (Incentive) & Circumstances dictate & 7.00 (6.17) \\
    & Selfless sacrifices & \\
    & Humilty & \\
    & President Obama & \\
    & EquityMarketReport & \\
\inlinegraphics{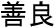} (Kindness)    & Really hateful lemmings & 7.81 (6.84) \\
    & Loving & \\
    & Dad & \\
    & Flowering orchids & \\
    & Dear robin & \\
\inlinegraphics{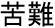}  (Misery) & Violence begets violence & 1.93 (3.91) \\
    & Indignations & \\
    & Sufferings    & \\
    & Sincerely & \\
    & Unconfessed sin &                                                                                    
\end{tabular}
}
\caption{Four poorly translated Chinese words, their accompanying true English translation, the five nearest English words to the translated vector, and the true (English) valence with the accompanying predicted valence of the translated vector.}
\label{table:fine-examples}
\end{table}

We explore the effect of poor translation on fine-grained sentiment analysis by taking a look at a few concrete examples of poor translations (Table \ref{table:fine-examples}). Table \ref{table:fine-examples} presents four different Chinese words with predicted valence of varying accuracy. We can see that even with poor translation ({\em i.e.}, the closest five words are all completely unrelated, as is the case with \textit{hungry}), sentiment is still accurately predicted. On the other hand we also show that it is possible to have related words but have poor prediction, as is the case with \textit{misery}, because the base sentiment predictor itself is not perfectly accurate. This suggests that the sentiment vector space, a hypothetical space produced by finding the sentiment value of each point in the original embedding space, which we term topological sentiment map, has two properties: i) it is maintained through linear transformation, and ii) it is ``flat" enough that highly accurate mapping (read: translation) between languages is not required to arrive at usable sentiment classification. This second property allows for the sentiment analysis of languages where a large amount of labeled material is not available at an extremely low cost, and can also be used to aid in many cross-lingual sentiment related tasks.

In the future we hope to extend both the analysis and experiments discussed here to other languages and applications. For example the cross-lingual fine-grained sentiment analysis techniques could possibly be used to study the change in sentiment of words in a single language over time, leading to new insights or re-affirming old ones. Future analysis could compare different transformations and their effect on sentiment analysis. The ability to produce a ``stable" topographic sentiment map could also be used to evaluate algorithms which create the vector spaces as well.

Lastly, future work will focus on developing other low-cost approaches, possibly by implementing \citet{Duong:16}’s technique to improve the accuracy and precision of sentiment regression. This would serve to demonstrate the (expected) limits of the linear transformation model’s in handling polysemy and subsequently its impact on the topological mapping of sentiment in vector space.

\section*{Acknowledgements}
We would like to thank Moustafa Abdalla for his help and discussions. The work was financially supported by the Natural Sciences and Engineering Research Council of Canada.

\bibliography{ijcnlp2017}
\bibliographystyle{ijcnlp2017}

\end{document}